\documentclass[letterpaper]{article}
\PassOptionsToPackage{table}{xcolor}
\usepackage[preprint]{aaai2027}
\usepackage[hyphens]{url}
\usepackage{graphicx}
\usepackage{natbib}
\usepackage{caption}
\usepackage{subcaption}
\usepackage{multirow}
\usepackage{microtype}
\usepackage{inconsolata}
\usepackage{booktabs}
\usepackage{amsmath}
\usepackage{amssymb}
\usepackage{makecell}
\usepackage{xcolor}
\DeclareCaptionStyle{ruled}{labelfont=normalfont,labelsep=colon,strut=off}

\newcommand{\protonet}{\textsc{Precog-RM}}
\newcommand{\auroc}{AUROC}
\newcommand{\qwenbig}{Qwen3-Next-80B-Thinking }
\newcommand{\qwensmall}{Qwen3-8B }
\newcommand{\dsmodel}{DeepSeek-V3.2 }
\newcommand{\oaimodel}{GPT-5-mini }
\newcommand{\embmodel}{\texttt{text-embedding-3-large }}
\newcommand{\usersim}{\texttt{gpt-4o-mini }}
\newcommand{\ood}[1]{\cellcolor{black!12}#1}

\title{Divergent Paths: Learning an Outcome-Predictive Representation of LLM Agent Trajectories}
\author{
    Alexander Gill\textsuperscript{\rm 1,\rm 2},
    Guram Gogia\textsuperscript{\rm 1},
    Francisco Guzm\'an\textsuperscript{\rm 1}
}
\affiliations{
    \textsuperscript{\rm 1}Handshake AI, \textsuperscript{\rm 2}University of Utah\\
    alex.gill@utah.edu
}

\begin{document}

\maketitle

\begin{abstract}
In this work, we reframe agentic trajectories as paths through a shared embedding space of actions and environmental observations. We show that successful and failed rollouts of the same task separate in this space, with simple per-task discriminators predicting held-out rollout outcomes at
up to $0.95$ AUROC across three agent benchmarks and three policy models. To exploit this signal across tasks, we introduce \protonet{}, a transformer-based model that meta-learns a representation in which outcome prediction can be performed in a few-shot setting. We show that \protonet{} generalizes to unseen tasks and models, and gives confident predictions of trajectory success long before termination. Deployed as an early stopping controller for Group Relative Policy Optimization (GRPO) on Tau2-Bench, \protonet{} cuts training FLOPs by up to 47.3\% while maintaining performance comparable to baseline GRPO.
\end{abstract}

\begin{links}
    \link{Code}{https://github.com/alexgill321/precog-rm}
\end{links}
 \begin{figure*}[!t]
  \centering
   \includegraphics[width=0.9\linewidth]{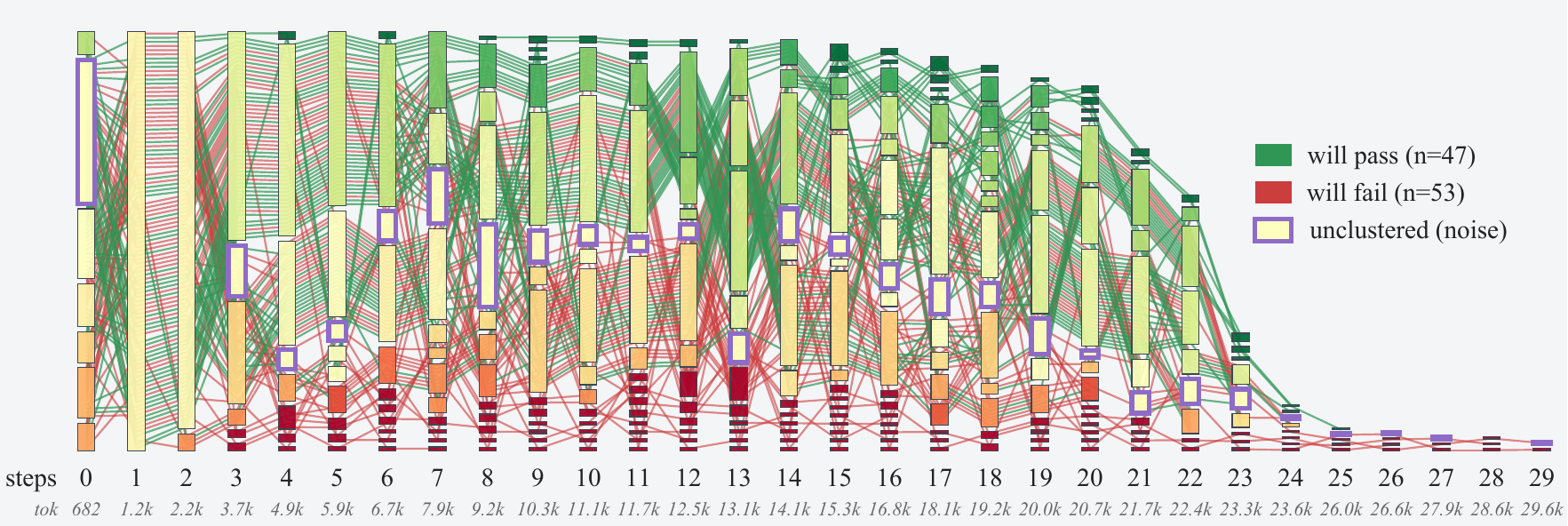}
   \caption{%
   Trajectory flow through paired action--response state clusters for 100 \texttt{retail/103} rollouts from Tau2-Bench. HDBSCAN over joint action--response embeddings yields one cluster assignment per rollout step; the first 30 steps are shown. Bar height encodes cluster occupancy, and bar color encodes empirical success probability (red $=0$, yellow $\approx 0.5$, green $=1$). Within each step, clusters are ordered by success rate from failure-associated (bottom) to success-associated (top). Thin lines trace individual rollouts between steps, colored by final outcome. Grey bars with purple outlines denote states HDBSCAN designates as noise. The bottom axis gives rollout step and mean cumulative tokens generated by that step. For more information on this diagrams construction, see Appendix~\ref{app:cluster-sep}.
 }
 \label{fig:trajectory-flow}
 \end{figure*}

\section{Introduction}
\label{sec:intro}

The shift from language modeling to tool-using agentic systems has expanded the medium LLMs must operate over. An agentic trajectory is made up of a sequence of interactions, each pairing an action taken by the agent with an observation returned by the environment \cite{yao2022react}, often spanning 10--100+ turns and 100K--500K+ tokens \cite{merrill2026terminal,luo2025ultrahorizon, vidgen2026apex}. 

This structure has made agentic trajectories the central object of a growing set of problems such as credit assignment \citep{feng2026group, he2026hierarchy, zhang2026reasoning}, failure diagnosis \citep{barke2026agentrx, tayebati2026tracer, ou-etal-2025-agentdiagnose}, and data filtering \citep{yang2026topo, lin-etal-2026-toolprm}. Most methods read the raw token sequence, where the tokens that determine an outcome are a small fraction of the surface form that carries them. Workarounds either prompt an LLM to analyze the trajectory \citep{zhuge2024agent, ou-etal-2025-agentdiagnose}, inheriting the cost and failure modes of the system under study, or compress trajectories into hand-crafted features \citep{tayebati2026tracer}, deciding in advance which properties matter. Neither transfers beyond the problem and domain it was built for.

Recent works instead treat actions \citep{feng2026group, he2026hierarchy} or action--observation pairs \citep{yang2026topo} as states recurring across multi-trial rollouts, so that a rollout becomes a \textbf{trajectory through a space of action-observation states}. Collapsing the many token realizations of an action or observation into shared states strips away the surface variation that obscures analysis signals. Unlike a fixed feature set or a prompted judge, the representation itself can be \textbf{learned}, allowing a single model to serve these problems across tasks and domains. A representation only matters, however, if it retains the signal these problems depend on. What credit assignment, failure diagnosis and data filtering all share is that each must tie what happened in a trajectory to how it ended. A representation that loses the outcome serves none of them.

We ask how much of a trajectory's outcome is recoverable from its actions and observations alone. We embed each action and observation with a frozen text encoder, a space we refer to as the action--observation space (AOS), and test whether same-task success and failure rollouts separate within it. We find that they do. Across three agentic benchmarks and three models, simple discriminators fit per-task on a handful of labeled rollouts predict held-out rollout outcomes at up to $0.95$ \auroc{}. Figure~\ref{fig:trajectory-flow}
shows the effect directly, with success and failure trajectories separating as their histories grow. 

Per-task fitting, however, transfers nothing, and its accuracy is bounded by the number of labeled rollouts a single task provides. We instead pose outcome prediction as a representation learning problem: learn an embedding of interaction histories under which same-task rollouts heading toward success separate from those heading toward failure, relating trajectories by outcome signal similarity rather than surface similarity. We introduce \protonet{}, a few-shot meta-learning model inspired by Prototypical Networks \citep{snell2017prototypical} which learns this embedding across tasks. On a new task, a single batch of labeled rollouts defines success and failure prototypes, and prediction reduces to distances against them, with no fitting at inference. \protonet{} separates better than per-task discriminators when labeled rollouts are
plentiful and increasingly outperforms them as they grow scarce, and it
transfers to unseen tasks and policy models. A variant predicts outcomes mid-rollout, abstaining when unsure. Finally, we deploy \protonet{} as an early stopping controller for GRPO \citep{shao2024deepseekmath} on $\tau^2$-Bench retail \citep{barres2025tau}: rollouts whose predicted success probability crosses the confidence threshold in either direction are terminated and assigned the predicted outcome as terminal reward, leaving the optimizer and advantage estimator untouched. The controller cuts training FLOPs by up to 47.3\% while keeping pass@1 within one standard deviation of the baseline.

We make four contributions: (i) a systematic study showing that same-task rollout outcomes are separable in embedded action--observation sequences, across three benchmarks and three policy models; (ii) \protonet{}, a meta-learned few-shot discriminator that carries this signal to unseen tasks and policies from a single batch of labeled rollouts; (iii) variants that predict outcomes mid-rollout and abstain when the signal is absent; and (iv) an end-to-end evaluation of \protonet{} as a GRPO rollout controller, cutting training
FLOPs by up to 47.3\% at near-baseline pass@1.

\section{Preliminaries}
\label{sec:preliminaries}

\subsection{Problem Setup}
\label{sec:objective}
We consider an LLM agent completing multi-step tasks from an instruction prompt $x$. At each step $t=1,\ldots,T$ the agent receives an observation $o_t$ and generates internal reasoning $r_t$ and a textual action $a_t$\footnote{We exclude $r_t$ from all representations: empirically it carries no signal complementary to $(a_t, o_t)$ (Appendix~\ref{app:signal-ablations}).}. A trajectory $\tau=(x,a_1, o_1, \ldots, a_T, o_T)$ terminates with a binary outcome $y \in \{0, 1\}$ indicating task success. The agent generates each action $a_t$ conditioned on the interaction history $h_t = (x, a_1, o_1, \ldots,a_{t-1}, o_{t-1})$. 

A single rollout is one realization of the stochastic process induced by $x$. On mixed-outcome tasks, identical starting conditions yield both successes and failures, so the outcome is decided along the path itself. We hypothesize that $h_t$ accumulates evidence of that decision as the trajectory unfolds. Our objective is a representation $\phi$ of trajectory histories under which same-task rollouts headed for success separate from those headed for failure, subject to the requirement that the success probability be trivially recoverable from $\phi$. Concretely, at any step $t$, we estimate the success probability of the in-progress trajectory from its history and a small same-task support set $S = \{(\tau^{(i)}, y^{(i)})\}_{i=1}^{|S|}$,
\begin{equation}
g\!\left(\phi(h_t),\, \phi(S)\right)
  \;\approx\; \Pr\!\left(y = 1 \mid h_t, S\right),
  \label{eq:objective}
\end{equation}
where $\phi(S) = \{(\phi(\tau^{(i)}), y^{(i)})\}_{i=1}^{|S|}$ denotes the embedded, labeled support set, and $g$ maps the embedded history and support to a probability of success. Approaches to Eq.~\ref{eq:objective} differ in where task adaptation occurs. Per-task discriminators hold $\phi$ fixed and fit $g$ on $S$ directly \cite{shen2025carl,wang2025information,feng2026group}, which is viable only insofar as a handful of rollouts can support the fit. We instead learn $\phi$ across tasks such that $g$ requires no fitting at all, placing the burden of the objective on the representation. Group-based RL training \citep{shao2024deepseekmath} supplies both structures of Eq.~\ref{eq:objective} for free: each completed group of $N$ same-task rollouts is a labeled support set $S$, and since advantages are computed from within-group statistics, uniform-outcome groups carry zero advantage, making mixed-outcome tasks precisely those on which a discriminator would be consulted.
\begin{figure*}[h]
    \centering
    \includegraphics[width=\textwidth]{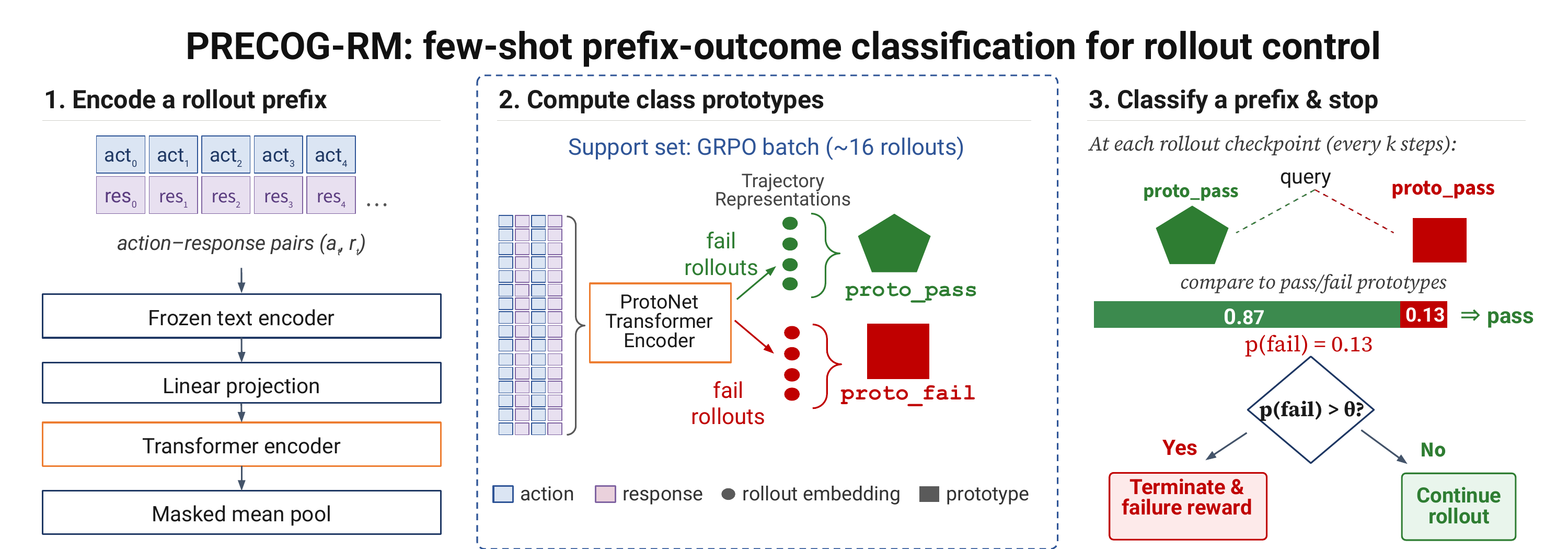}
   \caption{\textbf{\protonet{} architecture and deployment.} Action--observation pairs from a rollout prefix are embedded, linearly projected, and processed by a transformer encoder; masked mean pooling yields a trajectory representation $\vec{z}\in\mathbb{R}^{d'}$. A per-task support set $S$ defines pass/fail class prototypes $\vec{c}_{\mathrm{pass}}, \vec{c}_{\mathrm{fail}}$ as the mean embeddings of support rollouts. A query prefix is classified by softmax over negative Euclidean distances to the two prototypes; during generation, rollouts with $p(\mathrm{fail}) > \theta$ are terminated and assigned the failure reward.}
    \label{fig:architecture}
\end{figure*}
\subsection{Trajectory Embedding}
\label{sec:aos-rep}
Any instantiation of Eq.~\ref{eq:objective} must first decide how trajectories are encoded. A central hypothesis of this work is that the actions and environmental observations of an agentic trajectory carry most of its outcome-level signal. A sufficiently good embedding compresses each action or observation into a single state without losing outcome signal. We therefore embed each action $a_t$ and environmental response $o_t$ with a frozen text encoder, producing $\mathbf{e}_t^a, \mathbf{e}_t^o \in \mathbb{R}^d$. A trajectory becomes a path through this space, $\mathrm{AOS}(\tau)=\big(\mathbf{e}_1^a, \mathbf{e}_1^o, \ldots, \mathbf{e}_T^a, \mathbf{e}_T^o\big) \in \mathbb{R}^{2T\times d}$, comparable across rollouts free of token-level wording variance. We refer to this space as the \emph{action--observation space} (AOS) purely as shorthand, it is a semantic embedding space restricted to actions and observations. Every representation $\phi$ in this work is built on it. 

\section{Method: \protonet{}}
\label{sec:method}
Per-task fits of $g$ are bounded by one task's labeled rollouts, in practice a single batch: nothing they learn transfers, and raw AOS paths ($2T$ steps in $\mathbb{R}^d$) are too high-dimensional to fit. \protonet{} instead learns $\phi$ across tasks. A lightweight transformer encoder modeled after Prototypical Networks \citep{snell2017prototypical}, it meta-learns an embedding space in which successful and failed rollouts of the same task separate: where the AOS relates actions and observations by semantic similarity, \protonet{} relates entire trajectories by outcome similarity. It consumes raw AOS paths directly, forms task-specific pass/fail prototypes at inference from a single labeled batch, and reduces $g$ to prototype distances with no fitting on the support. 
\begin{table*}[h]
\centering
\small
\begin{tabular}{ll c cc ccc c}
\toprule
& & & \multicolumn{2}{c}{\textbf{Pooled}} & \multicolumn{3}{c}{\textbf{Clustered}} & \textbf{Learned} \\
\cmidrule(lr){4-5}\cmidrule(lr){6-8}\cmidrule(lr){9-9}
\textbf{Dataset} & \textbf{Model} & \textbf{$N$} & \textbf{NC} & \textbf{RBF}
  & \textbf{FT $n{=}1$} & \textbf{FT $n{=}2$} & \textbf{FT $n{=}5$} & \textbf{\protonet{}} \\
\midrule
\multirow{3}{*}{\textsc{Tau2-Bench}}
  & \textsc{DS-V3.2}\textsuperscript{$\star$}   & 28 & .901\tiny{±.008} & .928\tiny{±.009} & .860\tiny{±.014} & .868\tiny{±.013} & .852\tiny{±.016} & \textbf{.950}\tiny{±.009}\\
  & \textsc{Qwen3-80B}\textsuperscript{\dag}    & 38 & .862\tiny{±.013} & .911\tiny{±.010} & .917\tiny{±.012} & .917\tiny{±.009} & .884\tiny{±.010} & \textbf{.930}\tiny{±.013}\\
  & \textsc{GPT-5-mini}\textsuperscript{\ddag}  & 41 & .851\tiny{±.014} & .891\tiny{±.009} & .919\tiny{±.007} & .917\tiny{±.011} & .891\tiny{±.007} & \textbf{.925}\tiny{±.003}\\
\midrule
\multirow{3}{*}{\textsc{BFCLv3}}
  & \textsc{DS-V3.2}\textsuperscript{$\star$}   & 50 & .787\tiny{±.017} & .830\tiny{±.024} & .857\tiny{±.010} & .854\tiny{±.004} & .816\tiny{±.010} & \textbf{.876}\tiny{±.012}\\
  & \textsc{Qwen3-80B}\textsuperscript{\dag}    & 40 & .887\tiny{±.010} & \textbf{.926}\tiny{±.012} & .908\tiny{±.016} & .915\tiny{±.022} & .912\tiny{±.036} & .923\tiny{±.007}\\
  & \textsc{GPT-5-mini}\textsuperscript{\ddag}  & 20 & .853\tiny{±.005} & .927\tiny{±.011} & .913\tiny{±.001} & \textbf{.948}\tiny{±.015} & .942\tiny{±.017} & .873\tiny{±.020}\\
\midrule
\multirow{3}{*}{\textsc{AgentClinic}}
  & \textsc{DS-V3.2}\textsuperscript{$\star$}   & 14 & .744\tiny{±.013} & .773\tiny{±.012} & .688\tiny{±.009} & .676\tiny{±.015} & .680\tiny{±.021} & \textbf{.777}\tiny{±.037}\\
  & \textsc{Qwen3-80B}\textsuperscript{\dag}    & 13 & .785\tiny{±.021} & .849\tiny{±.014} & .792\tiny{±.018} & .782\tiny{±.004} & .744\tiny{±.013} & \textbf{.865}\tiny{±.029}\\
  & \textsc{GPT-5-mini}\textsuperscript{\ddag}  & 13 & .763\tiny{±.008} & .772\tiny{±.015} & .768\tiny{±.018} & .735\tiny{±.027} & .688\tiny{±.042} & \textbf{.781}\tiny{±.021}\\
\bottomrule
\end{tabular}
\caption{Outcome separability on held-out test tasks (\auroc{}, mean $\pm$ std across folds). Within each task, stratified 3-fold CV over its rollouts: the per-task discriminators (NC, RBF on the pooled representation; fastText on clustered state-ID sequences with $n$-gram order $n$) are fit on each train fold ($\sim$33 rollouts), while \protonet{} (per-benchmark trained on all three models' rollouts, Appendix~\ref{app:pools}) uses the same train fold
as its support set, with no per-task fitting. Folds are identical across all methods; \protonet{} is additionally averaged over 3 training seeds. $N$ = held-out test tasks per cell. Best per row in bold. \textsuperscript{$\star$}\,\texttt{deepseek-v3.2}; \textsuperscript{\dag}\,\texttt{qwen3-next-80b-a3b-thinking};
\textsuperscript{\ddag}\,\texttt{gpt-5-mini-2025-08-07}}
\label{tab:separability}
\end{table*}

\subsection{Architecture}
\label{sec:architecture}

\paragraph{Trajectory encoding.}
A learned linear projection maps each step embedding in $\mathrm{AOS}(\tau)$ to $\mathbb{R}^{d'}$ ($d'{=}128$), a bidirectional transformer encoder ($L{=}2$ layers,
$H{=}4$ heads) contextualizes the projected sequence, and masked mean pooling yields a fixed-length trajectory representation $\mathbf{z} \in \mathbb{R}^{d'}$. Appendix~\ref{app:ablations} reports ablations over the encoder design.

\paragraph{Prototypical classification.}
Given a few-shot batch $S$ of labeled rollouts from a task, class prototypes are computed as mean representations, $\mathbf{c}_k = \tfrac{1}{|S_k|}\sum_{i \in S_k} \mathbf{z}_i$
for $k \in \{0,1\}$ (fail/pass), where $S_k \subseteq S$ contains the support rollouts of class $k$, and $\mathbf{z}_i$ is the representation of $\tau^{(i)}$. Following \citet{snell2017prototypical}, a query trajectory $\tau_q$ is classified by a
softmax over negative distances to the prototypes: $
    \hat{p}(y = k \mid \tau_q) =
    \frac{\exp\big(-d(\mathbf{z}_q, \mathbf{c}_k)\big)}
         {\sum_{k'} \exp\big(-d(\mathbf{z}_q, \mathbf{c}_{k'})\big)},$
where $d(\mathbf{z}, \mathbf{z}') = \lVert \mathbf{z} - \mathbf{z}' \rVert_2$ is the Euclidean distance. In the terms of
Eq.~\ref{eq:objective}, the projection, transformer, and pooling constitute $\phi$, whereas $\mathbf{c}_k$ and $\hat{p}$ are the fixed readouts $g$: prototypes and distances involve no fitting on $S$.

\subsection{Training \& Inference}
\label{sec:training}
\paragraph{Training.}
\protonet{} is trained offline via episodic meta-learning, each episode mimicking the few-shot condition at inference: an episode samples one task uniformly from the training pool, draws disjoint support $S$ and query $Q$ sets from its rollouts ($N_S$ and $N_Q$ per class), and minimizes the negative log-likelihood of the query labels under the prototype-distance softmax, $\mathcal{L}_{\text{episode}} = -\frac{1}{|Q|} \sum_{(\tau_q, y_q) \in Q} \log \hat{p}(y = y_q \mid \tau_q)$; gradients flow through the projection and transformer encoder only, and the AOS is frozen. At inference, a single batch of labeled rollouts, e.g.\ a completed group from the RL loop, serves as $S$, and every subsequent rollout is scored by $\hat{p}$ in one forward pass. Because the encoder consumes AOS paths of any length, the same procedure scores an in-progress history $h_t$, enabling mid-rollout prediction.

\section{Experiments}
\label{sec:experiments}
\subsection{Experimental Setup}
\label{sec:exp-setup}
\paragraph{Benchmarks.}
We use three long-horizon agent benchmarks selected for their coverage of agent use cases across multiple domains. $\tau^2$-Bench~\citep{barres2025tau} is a customer service agent benchmark spanning the airline, retail, and telecom domains. BFCLv3~\citep{patil2025bfcl}, the Berkeley Function Calling Leaderboard multi-turn evaluation, covers base function calling, long-context, missing function, and missing parameter scenarios. AgentClinic~\citep{schmidgall2024agentclinic} evaluates tool-using clinical AI agents in the medical domain.

For all benchmarks, we screen each task with a 10-rollout sample and discard degenerate tasks, those in which every rollout succeeds or every rollout fails. Without both outcome classes, within-task separability is undefined; and since degenerate groups carry zero advantage in group-based RL (Section~\ref{sec:preliminaries}), mixed-outcome tasks are precisely those on which a discriminator would ever be consulted. For each surviving task, we generate 50 rollouts per model, embedding every action and observation into the
AOS with \embmodel{} ($d{=}3072$), the frozen text encoder of
Section~\ref{sec:aos-rep}. AgentClinic requires one further filter: its graded diagnosis can often be reached from internal knowledge alone, letting a discriminator separate outcomes without reading the trajectory. We therefore retain only tasks whose success rate drops under a no-tools ablation. Curation and action--observation extraction details for each benchmark are in Appendix~\ref{app:exp-setup-details}.

\paragraph{Models.} To test whether our findings hold across model capabilities, we generate rollouts with three models from three families, stratified by reported performance on Tau2-Bench Telecom\footnote{\url{https://artificialanalysis.ai/evaluations/tau2-bench}}: \dsmodel{} (78.9\%, strong), \qwenbig{} (41.5\%, mixed), and \oaimodel{}(25.1\%, weak).

\paragraph{Splits and training.} Within each benchmark, one third of tasks are held out as a test split shared across all models, so cross-model transfer is always evaluated on unseen tasks. The remaining tasks form the training pools, filtered to tasks with at least 8 rollouts of each class to support episodes with $N_S{=}N_Q{=}4$. To measure how generalization scales with training diversity, we train one \protonet{} per pool of (benchmark, model) rollouts, 25 pools in total (Appendix~\ref{app:pools}), with 3 seeds each. Every model trains for a fixed budget of 10{,}000 episodes with no early stopping or checkpoint selection. Hyperparameters are in Appendix~\ref{app:hyperparameters}, design ablations in Appendix~\ref{app:arch-ablations}.

\paragraph{Evaluation protocol.} To evaluate on a held-out task, we draw a stratified support set of $k$ rollouts per class ($k{=}4$ unless stated) and score the remaining rollouts, repeated over 5 seeded draws with identical draws for every method. We report \auroc{} averaged over draws, then tasks, then seeds.

\paragraph{Baselines.} In addition to \protonet{}, we use two baseline representations over the AOS. \textbf{Pooled} mean-pools $\mathrm{AOS}(\tau)$ into a single vector,
discarding step order, and \textbf{clustered} groups per-step embeddings into $K$ per-task interaction states, collapsing a trajectory to an order-preserving state-ID path (Appendix~\ref{app:cluster-sep}). Figure~\ref{fig:trajectory-flow} shows an example of a clustered task. Both are fixed, non-learned choices of $\phi$: all task adaptation happens in $g$, fit on each task's support. We fit three discriminators of increasing expressive power. 
\begin{itemize}
    \item \textbf{Nearest-Centroid (NC)}, on the pooled $\phi$: scores a rollout by the difference of its cosine similarities to the pass and fail class means.
    \item \textbf{RBF-SVM}, on the pooled $\phi$: fits a nonlinear max-margin boundary over the pooled vector, admitting curved decision boundaries in the full AOS feature space.
    \item \textbf{fastText}~\citep{joulin2017bag}, on the clustered $\phi$: a linear $n$-gram classifier over state-ID sequences, swept from an order-free bag of states ($n{=}1$) to longer subsequences ($n{=}2, 5$).
\end{itemize}
Together they bound what per-task fitting extracts from the AOS and set the reference we compare \protonet{} against.

\begin{figure}[t]
    \centering
    \includegraphics[width=0.9\columnwidth]{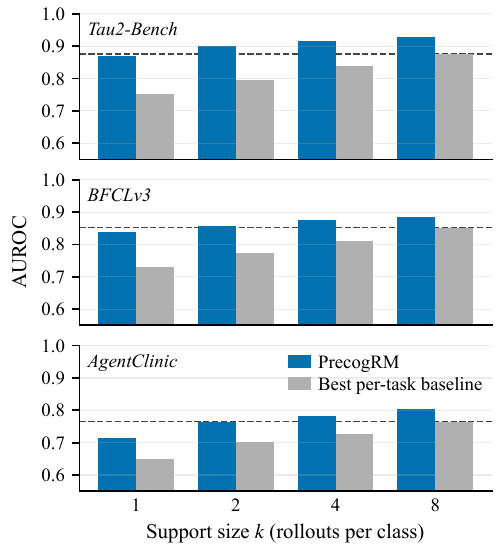}
    \caption{Few-shot discrimination on held-out tasks vs.\ support size $k$ per class. \auroc{} of \protonet{} (per-benchmark specialist, mean over 3 seeds) and the strongest per-task baseline at each support size $k$, on a fixed task cohort with identical seeded support draws. The dashed line marks the best baseline at $k{=}8$: \protonet{} matches it with $k{=}2$, a four-fold reduction in labeled rollouts. Per-method results in Appendix~\ref{app:support-set-full}.}
    \label{fig:fewshot}
\end{figure}

\subsection{Full-Support Discrimination}
\label{sec:exp-sep}
All evaluations in this section are on the held-out test tasks of
Section~\ref{sec:exp-setup}, unseen during \protonet{} meta-training. On these tasks, we evaluate \protonet{} against the per-task baselines at maximum per-task signal. Baselines are fit on a fold ($\sim$33 rollouts) of each test task and \protonet{} receives the same fold as its support set. \protonet{} here is the per-benchmark specialist: one encoder per benchmark, meta-trained on that benchmark's training tasks with rollouts from all three policies pooled. Table~\ref{tab:separability} reports per-task \auroc{} under stratified 3-fold cross-validation.

The baseline columns establish the separability properties of the space
itself. Outcomes are separable in every condition (best per-task
discriminator: $0.77$--$0.95$ \auroc{}), supporting the hypothesis that embedded actions and observations retain outcome-level signal. Separability is not uniform, as ceilings shift by $\sim\!0.12$ \auroc{} across benchmarks and $\sim\!0.05$ across models, with AgentClinic the consistent low performer (Appendix~\ref{app:agentclinic}).

Against this ceiling, \protonet{} matches or exceeds the best per-task fit in eight of nine conditions, taking the best score in seven. The exception, BFCLv3/\oaimodel{} ($.873$ vs.\ $.948$), is the only condition where order-sensitive discriminators win, a signal our position-free encoder ignores by design (Appendix~\ref{app:ablations}). 

\subsection{Few-Shot Discrimination on Held-Out Tasks}
\label{sec:exp-fewshot}
Deployment offers far less than full support, so we next evaluate outcome separability under a few-shot setting. Figure~\ref{fig:fewshot} compares \protonet{} against the strongest per-task baseline at $k \in \{1,2,4,8\}$ support rollouts per class, using identical seeded support draws on a fixed task set. \protonet{} improves on the strongest baseline at every support size on every benchmark. Its margin is largest at $k{=}1$ ($+0.06$--$0.12$ AUROC). At $k{=}2$, it matches or exceeds the strongest baselines at $k{=}8$, a four-fold reduction in the labeled rollouts required. 

\begin{table}[t]
\centering
\small
\setlength{\tabcolsep}{5pt}
\begin{tabular}{l ccc}
\toprule
& \multicolumn{3}{c}{\textbf{Evaluated on}} \\
\cmidrule(lr){2-4}
\textbf{Trained on} & \textsc{DS-V3.2} & \textsc{Qwen3-80B} & \textsc{GPT-5-mini} \\
\midrule
\textsc{DS-V3.2}    & .827        & \ood{.814}  & \ood{.763} \\
\textsc{Qwen3-80B}  & \ood{.768}  & .886        & \ood{.826} \\
\textsc{GPT-5-mini} & \ood{.776}  & \ood{.820}  & .861 \\
\addlinespace[2pt]
DS+Qwen             & .844        & .892        & \ood{.837} \\
DS+GPT              & .841        & \ood{.868}  & .874 \\
Qwen+GPT            & \ood{.801}  & .885        & .868 \\
\addlinespace[2pt]
All models          & .840        & .893        & .863 \\
\bottomrule
\end{tabular}
\caption{Cross-model generalization: held-out-task \auroc{} by training-pool
model composition (rows; within-benchmark pools, aggregated over benchmarks
and 3 seeds) and evaluated policy (columns). Shaded cells are
out-of-distribution (evaluated policy not in the pool).}
\label{tab:generalization}
\end{table}

\subsection{Cross-Model and Cross-Benchmark Generalization}
\label{sec:exp-generalization}
We next ask what the learned space generalizes over. Can rollouts from policies or domains unseen in meta-training still be separated? Table~\ref{tab:generalization} varies training-pool composition along the policy axis. We call an evaluation \emph{out-of-distribution} (OOD) when the policy or benchmark is not represented in the training pool, and \emph{in-distribution} (ID) otherwise.

Every model added to the training pool improves OOD performance on the held-out model. The shaded OOD cells rise from $0.795$ (mean, single-model pools) to $0.835$ (two-model pools), approaching the ID reference ($\approx\!0.86$). Since all models' rollouts share the same underlying tasks, an added model contributes no new tasks, and the gain is attributable to model diversity alone. The benchmark axis behaves oppositely (Appendix~\ref{app:cross-benchmark}). Unseen benchmarks fall to or below the \auroc{} of the untrained-encoder regardless of pool size. The all-benchmark trained \protonet{} matches the performance of domain specific trained counterparts, indicating that the learned space is domain-bound, but not sensitive to domain diversity in training.

\subsection{Prefix Prediction}
\label{sec:exp-prefix}

Above, \protonet{} predicts success for complete rollouts. Many applications instead call for predicting the outcome of a trajectory \emph{in progress}. We train a prefix-conditioned variant of \protonet{} per benchmark (all three models). Support trajectories are encoded at full length, but each training query is truncated to a uniformly sampled prefix $l \sim \mathrm{U}\{1, \ldots, T_q\}$ before encoding.

\begin{figure}[t]
\centering
\includegraphics[width=0.9\columnwidth]{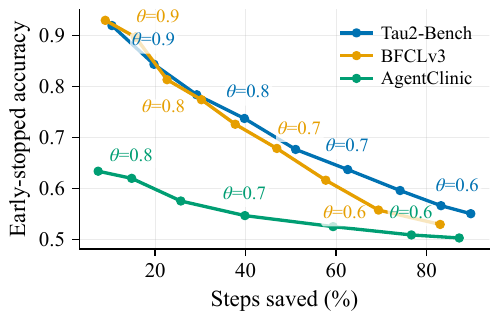}
\caption{Confidence-gated early stopping on held-out tasks: accuracy among
early-terminated rollouts vs.\ step-weighted compute saved, sweeping the confidence threshold $\theta$ (annotated). Each curve is the prefix-trained \protonet{} for that benchmark, evaluated in-domain. Thresholds at which
fewer than 10\% of rollouts terminate early are omitted.}
\label{fig:prefix-gated}
\end{figure}

Different tasks naturally have decisive actions or observations at different stages in their trajectories. A useful prefix predictor must abstain when outcome signal is not yet present. We therefore evaluate confidence-gated early stopping. Using predicted class probabilities, we early stop a prefix as a predicted success if $\hat{p} > \theta$ and a predicted failure if $\hat{p} < 1{-}\theta$. Rollouts which never cross $\theta$ run to completion. Figure~\ref{fig:prefix-gated} sweeps $\theta \in [0.55, 0.95]$, plotting accuracy on the stopped rollouts against the fraction of generation steps saved. On Tau2-Bench and BFCLv3, gating is effective: at $\theta{=}0.9$, $45\%$/$37\%$ of rollouts stop early with $0.84$/$0.90$ accuracy, saving $20\%$/$16\%$ of all generation steps. Relaxing to $\theta{=}0.8$ roughly doubles the savings at $0.74$/$0.77$ accuracy. AgentClinic performs near chance at all thresholds, but the model correctly abstains, stopping only $25\%$ of rollouts at $\theta{=}0.8$ compared to $69\%/56\%$ elsewhere.

\begin{figure*}[t]
\centering
\includegraphics[width=.9\textwidth]{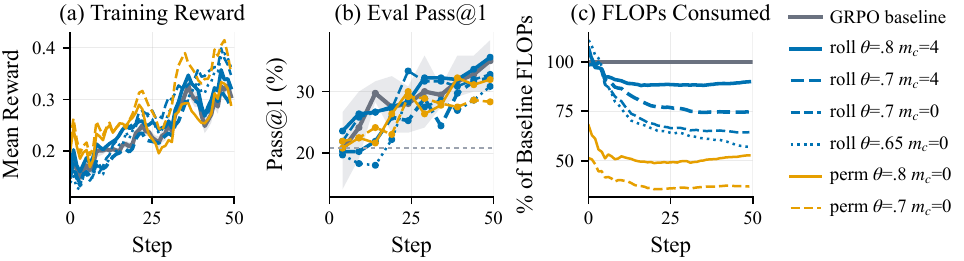}
\caption{(a) shows mean training reward per step; the gray band is the baseline's seed spread. (b) shows performance curves on the official test split; dashed gray is base Qwen3-8B performance (20.8\%). (c) shows cumulative FLOPs consumed as a percentage of baseline FLOPs at equivalent steps. All curves average 3 seeds. Color denotes support type, linestyle the threshold $\theta$, and linewidth $m_c$.}
\label{fig:training-curves}
\end{figure*}

\section{Application: Early Stopping in GRPO}
\label{sec:application}
Having established \protonet{} as an effective few-shot outcome predictor that can discriminate a trajectory's outcome before completion, we now explore its application on one of the motivations from Section~\ref{sec:intro}, RL post-training efficiency. In agentic RL, a single agentic task is often sampled eight or sixteen times, making rollout generation the primary compute bottleneck in the RL-training pipeline \cite{zhai2026selective,zhang2026train}. We therefore deploy \protonet{} to terminate confidently-predicted rollouts early, asking whether the saved generation compute can be realized \emph{without degrading the trained policy}.

\subsection{Setup}
\label{sec:grpo-setup}
We integrate a prefix-trained \protonet{} as a rollout controller inside the GRPO loop. For each new task encountered during training, the first batch of rollouts is generated to completion and used as the support set, defining the task's prototypes. Every subsequent rollout is scored after each action--observation pair. When the predicted success probability crosses a confidence threshold ($\hat{p} > \theta$ or $\hat{p} < 1{-}\theta$), the rollout is terminated early and assigned the predicted outcome as its reward. Rollouts that never cross $\theta$ run to natural termination and receive their true environment reward.

Three knobs govern the controller. The threshold $\theta$ trades savings against prediction risk. $m_c$ is the number of rollouts per group forced to run to completion: $m_c{=}4$ retains 25\% of each group as unbiased reward signal; $m_c{=}0$ allows all rollouts to stop. Per-task support is either taken only from the current run's completed rollouts (\textbf{Rolling}), or kick-started by a group of pre-collected rollouts (\textbf{Permanent}) generated from a different model (\qwenbig), exploiting the cross-model generalization found in Section~\ref{sec:exp-generalization}. Configurations are selected based on simulated early stopping calculations over baseline GRPO runs (see Appendix~\ref{app:gpro-details} for more details).

\paragraph{Protocol.} We train Qwen3-8B (LoRA, $r{=}32$) on the official $\tau^2$-Bench retail train split: 50 batches of 20 tasks, group size 16, three seeds per configuration, evaluating pass rate on the official test split. See Appendix~\ref{app:hyperparameters} for other training details. We report compute in FLOPS based on generation/training token counts and model parameters, and account for the embedding cost of scoring separately (Appendix~\ref{app:flops}).
\begin{table}[t]
\centering
\small
\setlength{\tabcolsep}{4pt}
\begin{tabular}{lcccc}
\toprule
Support & $\theta$ & $m_c$ & \makecell{FLOPs\\saved (\%)} & Pass@1 \\
\midrule
GRPO baseline & --- & --- & --- & 35.0\,\tiny{$\pm$\,3.6} \\
\midrule
rolling   & 0.8  & 4 & 10.1\,\tiny{$\pm$\,1.4} & 35.6\,\tiny{$\pm$\,2.1} \\
rolling   & 0.7  & 4 & 25.3\,\tiny{$\pm$\,3.6} & 32.2\,\tiny{$\pm$\,2.8} \\
rolling   & 0.7  & 0 & 35.5\,\tiny{$\pm$\,2.2} & 32.8\,\tiny{$\pm$\,6.1} \\
rolling   & 0.65 & 0 & 43.0\,\tiny{$\pm$\,4.2} & 30.8\,\tiny{$\pm$\,3.0} \\
\midrule
permanent & 0.8  & 0 & 47.3\,\tiny{$\pm$\,1.8} & 31.9\,\tiny{$\pm$\,3.7} \\
permanent & 0.7  & 0 & 63.0\,\tiny{$\pm$\,2.0} & 28.3\,\tiny{$\pm$\,3.4} \\
\bottomrule
\end{tabular}
\caption{GRPO on $\tau^2$-Bench retail with \protonet{} early stopping (50 batches,
Qwen3-8B; mean$\pm$std over 3 seeds). $\theta$: confidence threshold for early
termination; $m_c$: rollouts per group forced to run to completion. Support is
accrued from the current run (\emph{rolling}) or pre-collected per task
(\emph{permanent}). FLOPs accounting in Appendix~\ref{app:flops}. Pass@1 on the
official test split at matched training end.}
\label{tab:grpo}
\end{table}
\subsection{Results}
\label{sec:grpo-results}
Table~\ref{tab:grpo} and Figure~\ref{fig:training-curves} report 50-step GRPO runs. The most conservative configuration (rolling support, $\theta{=}0.8$, $m_c{=}4$) slightly exceeds baseline performance on the test set after training (pass@1 $35.6 \pm 2.1$ vs.\ $35.0 \pm 3.6$), while consuming $10.1\%$ less compute. All other configurations trade moderate losses in training gain for significant compute savings. Four of six configurations have final test set pass@1 mean values within one standard deviation of the baseline GRPO mean. Of these, permanent support, $\theta{=}0.8$, $m_c{=}0$ saves the most compute, at a $47.3\%$ reduction in training FLOPs. It has a mean pass@1 of $31.9$ on the test set, which retains $78\%$ of the post-training gain of baseline GRPO over default Qwen3-8B. Only permanent support, $\theta{=}0.7$, $m_c{=}0$ saves more compute ($63.0\%$), but its pass@1 of $28.3$ only retains $53\%$ of performance gain compared to baseline, indicating that this configuration crosses the point where prediction errors materially degrade the learning signal.

\paragraph{Training Trends.}
Training reward trajectories are consistent across configurations (Figure~\ref{fig:training-curves}a late-training reward $0.29$--$0.37$ vs.\ baseline $0.30$),  indicating that assigning predicted rewards to terminated rollouts does not bias the group-relative advantage signal. Compute savings differ by support strategy (Figure~\ref{fig:training-curves}c): permanent-support configurations save from the first step, since prototypes are available before any rollout is scored, whereas rolling support begins at or above baseline cost, paying the embedding overhead of building support sets, with savings beginning once tasks accumulate viable support. All configurations except rolling $\theta{=}0.65$ saturate in cumulative FLOPs saved by late training.

\section{Related Work}
\label{sec:related}

\paragraph{Representations of agent trajectories.}
Agentic trajectories are the central object of a growing set of analyses, including credit assignment \citep{feng2026group, he2026hierarchy, zhang2026reasoning}, failure diagnosis \citep{barke2026agentrx, tayebati2026tracer, ou-etal-2025-agentdiagnose}, and training-data curation \citep{yang2026topo}. These methods operate on raw token sequences, hand-crafted features, or prompted LLM judges, each designed for a single problem and committed in advance to the trajectory features that matter. Closest to our view, TopoCurate \citep{yang2026topo} merges action-observation pairs from multi-trial rollouts into a per-task interaction topology and uses it to curate training data. Related in spirit, \citet{sun-etal-2026-llm} recover correctness signals from the step-wise geometry of a model's hidden states. We adopt the state abstraction of the former as our entry point, but read the model-external action--observation interface rather than the policy's internals. We establish outcome separability as a general property of trajectories in this space and train a single encoder that carries it to unseen tasks and models.

\paragraph{Prefix-outcome prediction and process reward.}
Process reward models provide intermediate signal for incomplete generations through step-level annotation \citep{lightman2024lets}, Monte Carlo construction \citep{wang2024math}, or implicit rewards recovered from outcome labels \citep{yuan2024free}, extending outcome-level verification \citep{cobbe2021training, uesato2022solving}. This line targets mathematical reasoning and dense per-step reward; ToolPRM extends it to function calling, scoring intra-call decisions during structured generation \citep{lin-etal-2026-toolprm}. \protonet{} instead predicts a single terminal outcome from a multi-turn action-observation prefix, adapts to a new task from a small support set following prototypical networks \citep{snell2017prototypical}, and uses the prediction to gate rollout continuation, substituting the predicted terminal reward for stopped rollouts.

\paragraph{Rollout control in RL post-training.}
Classical RL truncates rollouts at fixed horizons and bootstraps the unobserved return with a learned critic \citep{sutton1988learning, schulman2015gae, mnih2016asynchronous}; time limits treated as terminal states are themselves a source of bias \citep{pardo2018time}. Recent LLM RL treats the rollout budget as an optimization target \citep{surana2026generate}. GReSO filters prompts before generation by expected usefulness \citep{zheng2026act}. AERO adapts rollout allocation to avoid zero-advantage groups \citep{zhang2026train}. Selective Rollout is closest to our setting: it terminates agent rollouts mid-trajectory using a prefix-convergence heuristic \citep{zhai2026selective}. Work revising the GRPO objective itself \citep{yu2026dapo, liu2025drgrpo, kazemnejad2024vineppo} is orthogonal. \protonet{} leaves the objective, advantage estimator, and clipping rule untouched and modifies only which rollouts run to completion.

\section{Conclusion}
We showed that an agentic rollout's outcome is recoverable from its embedded actions and observations: simple per-task discriminators separate same-task successes from failures across three benchmarks and three policy models. \protonet{} lifts this signal beyond the single task, meta-learning a representation in which outcome prediction reduces to distances against few-shot prototypes. It matches per-task fitting at full support, dominates it in the few-shot regime, and generalizes across policies. Gated by confidence, it serves as an early stopping predictor, converting predictions into compute savings. For GRPO on $\tau^2$-Bench retail, \protonet{} cuts training FLOPs by up to 47.3\% with pass@1 within one standard deviation of the baseline, leaving the optimizer untouched. We see outcome-predictive and other learned trajectory representations as reusable primitives for future work in agentic trajectory analysis. 

\bibliography{references}
\clearpage

\appendix
\section{Details On Clustering in the Action-Observation Space}
\label{app:cluster-sep}

The \textbf{clustered} representation of Section~\ref{sec:exp-setup} abstracts agentic trajectories one step beyond the raw AOS. Embedded actions and observations are clustered by similarity into discrete per-task states, following the topology construction of \citet{yang2026topo}. Each task becomes a graph of interaction states connected by the transitions agents take between
them, and a trajectory becomes a path through this graph. In the terms of Eq.~\ref{eq:objective}, the state-ID map is a second fixed, non-learned choice of $\phi$, coarser than mean pooling but order-preserving. The discriminators below are per-task choices of $g$ fit on its output. This appendix details the construction and the discriminator family, of which the main text reports only
the strongest (fastText). 

\subsection{Topology Construction}
Given the $n \le 50$ rollouts $\{\tau^{(i)}\}_{i=1}^{n}$ of a single task, we form one vector per step by concatenating the action and observation embeddings of Section~\ref{sec:aos-rep},
$[\mathbf{e}_t^a \,\|\, \mathbf{e}_t^o] \in \mathbb{R}^{2d}$, and pool all steps of all rollouts. We cluster this pool with
HDBSCAN~\citep{mcinnes2017hdbscan} over cosine distances between the
row-normalized step vectors (\texttt{min\_cluster\_size}=5), yielding $K$ interaction states per task, with $K$ set by the algorithm rather than fixed in advance. Steps that HDBSCAN leaves unassigned are collected into a single additional outlier state, for $K{+}1$ states in total. The clustering defines the map $\phi$ taking each rollout to a state-ID sequence $\phi(\tau^{(i)})=(k_1^{(i)},\ldots, k_{T_i}^{(i)})$ with $k_t\in\{0,\ldots,K\}$. For AgentClinic we drop each trajectory's final step before clustering, as its closing diagnosis turn reveals the graded answer (Appendix~\ref{app:exp-setup-details}). Figure~\ref{fig:trajectory-flow} shows the resulting structure for one Tau2-Bench task, illustrating how pass and fail trajectories diverge as their interaction histories grow. 

\subsection{fastText over State-ID Sequences}
The clustered discriminator of Table~\ref{tab:separability} is a fastText classifier~\citep{joulin2017bag} trained per task on the state-ID sequences. Each sequence is treated as a document over the state vocabulary: every $n$-gram of states up to a maximum order receives a learned 20-dimensional embedding, the embeddings of all $n$-grams in a sequence are averaged, and a
linear softmax maps the average to a pass probability, giving a scalar score interpretable as $\hat{p}(\text{pass}\mid\tau)$ for each held-out rollout.

The maximum order $n$ controls how much sequential structure the
discriminator can express. At $n{=}1$ the model is an order-free bag of
states, sensitive only to which states a trajectory visits and how often; $n{=}2$ adds adjacent-state transitions; $n{=}5$ adds longer motifs. Sweeping $n \in \{1,2,5\}$ therefore traces how separability changes as progressively more of the path's order is made available, from pure state occupancy to multi-step subsequences.

\section{Signal Ablations}
\label{app:signal-ablations}

The AOS interleaves two channels: what the agent did ($a_t$) and what the environment returned ($o_t$). We ablate each channel under the Table~\ref{tab:separability} discriminator families, on the held-out test tasks of each (benchmark, model) cell (the Table~\ref{tab:separability} evaluation sets) with the same stratified
3-fold cross-validation, reporting means across folds. For each channel we report the \emph{raw ceiling}, the best of NC and RBF on that channel's pooled embedding, and the \emph{clustered ceiling}, the best fastText $n \in \{1, 2, 5\}$ over state-ID sequences re-clustered from that channel's own step embeddings (Appendix~\ref{app:cluster-sep}). 

All clusterings follow Appendix~\ref{app:cluster-sep}: for each task and channel, the step vectors of all rollouts are pooled, row-normalized, and clustered with HDBSCAN over cosine distances (\texttt{min\_cluster\_size}=5), with unassigned steps collected into one outlier state. The two full-trajectory columns read identical content but differ in construction. \emph{Paired} is the joint construction of Appendix~\ref{app:cluster-sep}. Its raw readout pools the same concatenated vectors (unit-normalized mean over steps). \emph{Indep.}\ instead runs HDBSCAN once on the action embeddings alone and once on the observation embeddings alone, producing two disjoint state vocabularies whose IDs are offset so they never collide. It then encodes a trajectory as the interleaved sequence $a_1, o_1, a_2, o_2, \dots$ over the union vocabulary. Its raw readout pools all $2T$ single-channel step embeddings in the shared $d$-dimensional space (unit-normalized mean). The single-channel columns have states clustered from that channel's own step embeddings, raw
readout from that channel's pooled vector.

\begin{table*}[t]
\centering
\small
\setlength{\tabcolsep}{5pt}
\begin{tabular}{ll c cccc cccc}
\toprule
& & & \multicolumn{4}{c}{Raw ceiling} & \multicolumn{4}{c}{Clustered ceiling} \\
\cmidrule(lr){4-7}\cmidrule(l){8-11}
Dataset & Model & $N$ & $a_t$ & $o_t$ & indep. & paired & $a_t$ & $o_t$ & indep. & paired \\
\midrule
\multirow{3}{*}{Tau2-Bench}
& GPT-5-mini & 41 & .889 & .870 & .882 & \textbf{.890} & .897 & .919 & .911 & \textbf{.921} \\
& Qwen3-80B  & 38 & \textbf{.910} & .878 & .907 & \textbf{.910} & .928 & .929 & \textbf{.936} & .915 \\
& DS-V3.2    & 28 & \textbf{.929} & .900 & .927 & .928 & .893 & \textbf{.955} & .952 & .866 \\
\midrule
\multirow{3}{*}{BFCLv3}
& GPT-5-mini & 20 & .922 & .896 & .921 & \textbf{.927} & .937 & .917 & .927 & \textbf{.948} \\
& Qwen3-80B  & 40 & .921 & .880 & .922 & \textbf{.925} & .920 & \textbf{.929} & .916 & .916 \\
& DS-V3.2    & 50 & .803 & .822 & .818 & \textbf{.830} & .811 & .855 & .855 & \textbf{.859} \\
\midrule
\multirow{3}{*}{AgentClinic}
& GPT-5-mini & 13 & .770 & .754 & .767 & \textbf{.772} & .664 & .708 & .689 & \textbf{.769} \\
& Qwen3-80B  & 13 & .831 & .842 & .832 & \textbf{.849} & .760 & .806 & \textbf{.809} & .795 \\
& DS-V3.2    & 14 & .725 & .759 & .763 & \textbf{.773} & .704 & .708 & \textbf{.744} & .693 \\
\bottomrule
\end{tabular}
\caption{Channel ablation: per-task \auroc{} ceilings (mean across 3 CV
folds, aggregated over each cell's $N$ held-out test tasks, the Table~1
evaluation sets) when the representation reads only actions ($a_t$),
only observations ($o_t$), or the full trajectory under disjoint
(indep.) or joint (paired) channel treatment. Raw = best of NC/RBF;
clustered = best fastText $n \in \{1,2,5\}$ on states clustered from
that channel alone. Best per row within each ceiling in bold.}
\label{tab:channel}
\end{table*}

Table~\ref{tab:channel} supports the constructions used throughout the
paper. Under the raw ceiling, paired carries the most signal --- best or tied on eight of nine cells --- and no single channel is reliably
sufficient. Hence we operate raw baselines (NC, RBF) on the paired pooled vector. Under clustering, paired again wins most often (four of nine cells, vs.\ three for disjoint vocabularies), so we operate clustered baselines using the joint construction of Appendix~\ref{app:cluster-sep}. \protonet{} itself consumes the interleaved step sequence rather than concatenated pairs. Comparing the single channels shows they are largely interchangeable:
either alone recovers most of the paired ceiling on every cell, and the
gap between them is small ($\leq 0.041$ raw) with a sign that flips by
benchmark. Actions edge out observations on five of six Tau2-Bench
and BFCLv3 cells, observations on AgentClinic.

\subsection{Excluding Internal Reasoning}
The problem setup of Section~2 excludes the agent's internal reasoning
$r_t$ from all trajectory representations. This subsection gives the
supporting evidence, under the exact protocol of the channel ablation
above: same evaluation set, folds, aggregation, and both discriminator
families. Of our three policies, only \qwenbig{} exposes its reasoning
traces, so the analysis covers the Tau2-Bench/\qwenbig{} cell --- 37 of
the 38 test tasks of Table~\ref{tab:separability} (one task lacks enough reasoning-aligned rollouts). Each step's reasoning text is embedded with the same frozen
encoder, giving three channels: \emph{AO} (the paired construction,
recomputed here as the reference), \emph{R} (reasoning steps alone), and \emph{AO$\oplus$R} (the concatenated $[a_t \| o_t \| r_t]$ step vector, mirroring the paired construction with the third channel added).

\begin{table}[t]
\centering
\small
\begin{tabular}{l cc}
\toprule
Channel & Raw ceiling & Clustered ceiling \\
\midrule
AO           & \textbf{.910} & \textbf{.916} \\
R            & .886          & .790 \\
AO$\oplus$R  & .906          & .852 \\
\bottomrule
\end{tabular}
\caption{Reasoning channel ablation on the Tau2-Bench/\qwenbig{}
evaluation set ($N{=}38$; per-task \auroc{}, mean across 3 CV folds;
protocol and ceilings as in Table~\ref{tab:channel}). AO = paired
action--observation; R = reasoning alone; AO$\oplus$R = concatenated
per-step vector. Best per column in bold.}
\label{tab:reasoning}
\end{table}

Reasoning alone is strongly predictive (raw ceiling $.886$), confirming
that the traces reflect trajectory state, but it carries no signal
complementary to the actions and observations: adding it leaves the raw
ceiling essentially unchanged ($.906$ vs.\ $.910$). Under clustering
the redundancy becomes a cost. Reasoning's clustered ceiling falls $0.096$ below its raw one, where AO loses nothing, and folding it into the joint clustering drags the ceiling
from $.916$ to $.852$. We therefore exclude $r_t$ throughout,
noting the scope of this evidence: one policy and one benchmark, as the
remaining policies do not expose their reasoning traces.

\section{Architecture Design Ablations}
\label{app:ablations}
We fixed \protonet{}'s configuration with a single-seed design study run before the final training sweep. The ablated encoder is trained on the Tau2-Bench/\qwenbig{} train-split tasks (4{,}000 episodes) and evaluated on held-out test tasks (Table~1 evaluation sets), under the
standard support protocol ($k{=}4$ per class, 5 seeded draws) in three
slices: \emph{within} (same cell), \emph{cross-model} (Tau2-Bench under
the other two policies), and \emph{cross-domain} (the six cells of the
other two benchmarks). This appendix ablates the encoder's inputs and
architecture (Table~\ref{tab:ablations}); the training-recipe and
readout ablations, under the same protocol, are in
Appendix~\ref{app:arch-ablations}. All values are mean per-task
\auroc{}.

Rows modify one choice at a time. The \emph{action/observation flag}
appends a binary input dimension to each interleaved step token, marking whether the token embeds an action or an observation, before the linear projection. \emph{Positional embeddings} add a learned per-position embedding (indexed by the token's position in the interleaved sequence) to the projected tokens before the transformer; the \emph{permuted steps} control re-evaluates that positional model with step order shuffled at test time, isolating whether it actually exploits order. \emph{No transformer} removes the encoder stack entirely ($L{=}0$), leaving the linear projection followed by masked mean pooling prototypes over a pooled linear map of the frozen step embeddings. The remaining rows vary depth $L$, projection width $d'$, dropout (on/off). \emph{Attention pooling} replaces the uniform masked mean with a learned single-query attention pool. Each encoded step is scored against a learned query vector and the trajectory embedding is the softmax-weighted sum, allowing the encoder to weight informative
steps rather than average them.

\begin{table}[t]
\centering
\small
\setlength{\tabcolsep}{5pt}
\begin{tabular}{l ccc}
\toprule
& \textbf{Within} & \textbf{X-model} & \textbf{X-domain} \\
\midrule
Full configuration        & \textbf{.906} & .886 & .687 \\
\midrule
+ action/obs.\ flag       & .902 & .891 & .675 \\
+ positional embeddings   & .814 & .785 & .661 \\
\quad + permuted steps    & .811 & .786 & .652 \\
No transformer            & .803 & .787 & .648 \\
$L{=}1$ layer             & .891 & .856 & \textbf{.716} \\
$L{=}4$ layers            & .895 & .889 & .659 \\
$d'{=}32$                 & .884 & .866 & .679 \\
$d'{=}256$                & .881 & .870 & .701 \\
No dropout                & .893 & \textbf{.895} & .689 \\
Attention pooling         & .901 & .894 & .694 \\
\bottomrule
\end{tabular}
\caption{Encoder architecture ablations (\auroc{}, mean over held-out
test tasks; single seed). Trained on Tau2-Bench/\qwenbig{}; Within =
same cell, X-model = Tau2-Bench under the other two policies, X-domain = mean over the six cells of the remaining benchmarks.}
\label{tab:ablations}
\end{table}

Two choices dominate: positional embeddings hurt ($-0.09$ within, and
the permutation control shows the positional model does not even use the order it is given: $.811$ shuffled vs.\ $.814$), and removing the
transformer costs $-0.10$. Both replicate when retrained on the mixed
pool of all nine cells (full $.852$; positions $.749$; no transformer
$.744$). Everything else is flat: no ablation improves within-cell
performance over the full configuration, and the few that improve
cross-model or cross-domain transfer do so only marginally and at a
within-cell cost (e.g., $L{=}1$: $+0.03$ cross-domain, $-0.015$ within). The final configuration is therefore the simplest consistent with the table: no positional embeddings, no type flag, $L{=}2$, $d'{=}128$, mean pooling.

\section{Dataset Additional Details}
\label{app:exp-setup-details}

For all benchmarks a rollout is a sequence of steps $(a_t, o_t)$, each step pairing an action text with the
observation text the environment returned for it. Every text is embedded
independently with \embmodel{} ($d{=}3072$); empty fields embed a
literal \texttt{<empty>} sentinel. For every (benchmark, model) cell we
screen each task with a 10-rollout sample, discard degenerate tasks (all rollouts succeed or all fail), and generate 50 rollouts per surviving task, each scored by the benchmark's own checker for a binary reward.
One third of each benchmark's tasks (54 Tau2-Bench, 60 BFCLv3, 64
AgentClinic) are held out as a test split shared across all three
policies; the remainder form the training pools of Appendix~\ref{app:pools}. Two class-count filters apply downstream of the split. Training pools
keep tasks with at least 8 rollouts of each outcome class (these
produce the training-pool counts of Table~\ref{tab:pool-sizes}).
Evaluation keeps test tasks with at least 5 rollouts of each class ---
the structural minimum for the $k{=}4$ support protocol, which draws 4
support rollouts per class and needs at least one of each class left
over to score. Intersected with the test split, this yields the
per-cell task counts $N$ of Table~\ref{tab:separability}. The subsections below give each benchmark's step definition and extraction rules.

\subsection{Tau2-Bench}

Tasks span the airline, retail, and telecom domains; the agent converses with the benchmark's simulated user (\usersim) and calls domain APIs, and a rollout succeeds when the final database state and required communications match the task's gold specification. The \qwenbig{} screen over the official task sets (50 airline, 114 retail, 114 telecom) retained 200 tasks: 39 airline, 85 retail, 76 telecom.
For each model, we then generate 50 rollouts on the same 200
tasks. At analysis time, 123 (\oaimodel), 111 (\qwenbig), and 94
(\dsmodel) of these have mixed outcomes under the respective policy.

A step is one assistant turn. The \emph{action} text concatenates the
assistant's message content with its serialized tool calls (function
name and arguments); turns with neither embed the \texttt{<empty>}
sentinel. The \emph{observation} is the environment text the next turn
conditions on: the tool execution result, or the user simulator's
reply. For the final step, which has no following turn, the observation
is the last user or tool message that followed the action.

\subsection{BFCLv3}
Multi-turn function-calling tasks from the base, missing-function,
missing-parameter, and long-context categories, scored by the
benchmark's state- and response-matching checker. From the 800-task
multi-turn suite, 400 tasks were trialed with the \qwenbig{} 10-rollout
screen; the 200 with mixed outcomes were retained (47 base, 41
long-context, 60 missing-function, 52 missing-parameter) and extended
to 50 rollouts. At analysis time, 56 (\oaimodel), 122 (\qwenbig), and
143 (\dsmodel) have mixed outcomes.

Unlike Tau2-Bench, where an assistant turn issues at most one API call,
a BFCLv3 agent may call several functions within a single assistant
step. Rollout logs are therefore parsed into $(a_t, o_t)$ steps at one
of two granularities. For \qwenbig{} and \dsmodel{}, each individual tool
call is its own step: the \emph{action} is the single call (function
name and arguments) and the \emph{observation} is that call's return
value. Assistant steps with text but no calls are kept as single steps whose observation is the next user prompt, and empty steps are dropped. For \oaimodel{}, steps instead follow the executor's turn/step
structure: all calls issued within one assistant step form a single
bundled action, paired with the concatenation of their return values. Every no-call step is kept as a turn boundary (the assistant's text or \texttt{<none>}, with the next user prompt as observation). The two granularities reflect the order in which the cells were embedded rather than
a design choice; on a held-out subset of 25 \qwenbig{} and 23 \dsmodel{} tasks, the two schemes agree within $0.01$ AUROC. All texts are truncated to 8{,}000 characters (within the encoder's 8{,}192-token limit).

\subsection{AgentClinic}

Scenarios derive from MedQA-Extended in which a doctor agent interviews a patient agent, may order tests from a measurement agent
(\texttt{REQUEST TEST}), and ends by committing to a diagnosis
(\texttt{DIAGNOSIS READY}), graded against the scenario's reference
diagnosis. AgentClinic is not pre-screened: all 214 MedQA-Extended
scenarios are rolled out 50 times by each policy, and degenerate
scenarios are filtered per model afterwards, leaving 150 (\oaimodel),
117 (\qwenbig), and 112 (\dsmodel) mixed-outcome scenarios.

A step is one doctor turn. The \emph{action} is reduced to the
canonical act the environment dispatches on: the question asked
(speaker prefix stripped), \texttt{REQUEST TEST: <test>}, stripping conversational preamble so the embedding reflects what was asked or ordered rather than the surrounding conversational context. The \emph{observation} is the patient's answer or the measurement result.

Two AgentClinic-specific controls apply downstream of extraction.
\emph{Leakage:} the closing diagnosis turn states the answer the grader
scores, so every representation (pooled, clustered, and \protonet{}
input) drops each trajectory's final step. \emph{Tool reliance:} we rerun every mixed-outcome scenario with a doctor prompt that removes the ability to request tests and keep, for evaluation, only scenarios whose success rate drops by more than a $0.1$ margin with statistical significance. We retain 26 of 150 scenarios for \oaimodel{}, 33 of 117 for \qwenbig{}, 43 of 112 for \dsmodel{}. All AgentClinic evaluations are restricted to each model's own tool-reliant scenarios, and meta-training pools retain all mixed-outcome scenarios (Appendix~\ref{app:pools}). Appendix~\ref{app:agentclinic} details the tool-ablation test and analyzes why this filter is needed.

\section{Training Pools}
\label{app:pools}

Every \protonet{} in this work is trained on a \emph{pool}: the union of
train-split rollouts from a chosen subset of the nine (benchmark, model)
cells. Varying pool composition along the model and benchmark axes yields 25
pools:

\begin{itemize}
    \item \textbf{Single-model pools} (9): one benchmark, one policy.
    \item \textbf{Two-model pools} (9): one benchmark, each pair of policies.
    \item \textbf{Benchmark specialists} (3): one benchmark, all three
    policies. These are the \protonet{} models of
    Table~\ref{tab:separability} and Figure~\ref{fig:fewshot}.
    \item \textbf{Two-benchmark pools} (3): each pair of benchmarks, all
    three policies.
    \item \textbf{Generalist} (1): all benchmarks, all policies.
\end{itemize}

Within a benchmark, all pools share the same underlying tasks, so adding a policy changes only whose rollouts are present, never which tasks are seen; this is what licenses attributing cross-model gains to policy diversity (Section~\ref{sec:exp-generalization}). Table~\ref{tab:pool-sizes} lists the train-split task counts per cell after the $\ge 8$ rollouts-per-class filter
of Section~\ref{sec:exp-setup}.

\begin{table}[t]
\centering
\small
\begin{tabular}{l ccc}
\toprule
& \textsc{DS-V3.2} & \textsc{Qwen3-80B} & \textsc{GPT-5-mini} \\
\midrule
Tau2-Bench  & 66 & 73 & 82 \\
BFCLv3      & 93 & 82 & 36 \\
AgentClinic & 71 & 79 & 99 \\
\bottomrule
\end{tabular}
\caption{Training tasks per (benchmark, model) cell after filtering to tasks with at least 8 rollouts of each class.}
\label{tab:pool-sizes}
\end{table}

\section{Training Ablations}
\label{app:arch-ablations}
Under the protocol of Appendix~\ref{app:ablations} we ablate the
training recipe and the inference rules around the fixed encoder.

Rows vary one choice at a time. The episode-budget rows scale
meta-training length ($0.2\times$ and $2\times$ the 4{,}000-episode
study budget); $N_S{=}N_Q$ sets the per-class support and query sizes
of each training episode. \emph{Prefix training} truncates each
episode's queries to random step prefixes, teaching the encoder to
embed partial rollouts --- the variant deployed for mid-rollout scoring
in Sections~\ref{sec:exp-prefix} and~\ref{sec:application}. \emph{Untrained encoder} evaluates a randomly
initialized encoder under the identical support protocol. The readout
rows re-score the trained checkpoint with alternative inference rules:
prototype distances in cosine rather than Euclidean geometry; $k$NN
($k{=}5$, mean label of the nearest support embeddings); logistic
regression fit on the support embeddings; and support drawn without
class stratification. Table~\ref{tab:design-ablations} reports mean
per-task \auroc{}.

\begin{table}[t]
\centering
\small
\setlength{\tabcolsep}{5pt}
\begin{tabular}{l ccc}
\toprule
& \textbf{Within} & \textbf{X-model} & \textbf{X-domain} \\
\midrule
Full configuration        & .906 & .886 & .687 \\
\midrule
\multicolumn{4}{l}{\emph{Training}} \\
$0.2\times$ episodes      & .891 & .885 & .716 \\
$2\times$ episodes        & .891 & .872 & .661 \\
$N_S{=}N_Q{=}1$           & .873 & .873 & .691 \\
$N_S{=}N_Q{=}8$           & .881 & .881 & .669 \\
Prefix training           & .879 & .863 & .708 \\
Untrained encoder         & .826 & .830 & .740 \\
\midrule
\multicolumn{4}{l}{\emph{Readout (evaluation-only)}} \\
Cosine distance           & .902 & .892 & .676 \\
$k$NN ($k{=}5$)           & .849 & .841 & .623 \\
Logistic regression       & .901 & .886 & .683 \\
Unstratified support      & .897 & .878 & .659 \\
\bottomrule
\end{tabular}
\caption{Training-recipe and readout ablations (\auroc{}, mean over
held-out test tasks; single seed; slices as in
Table~\ref{tab:ablations}). Readout rows re-score the study's base
checkpoint with alternative inference rules.}
\label{tab:design-ablations}
\end{table}

The training recipe is forgiving: the episode budget is flat from
$0.2\times$ to $2\times$; $N_S{=}N_Q{=}4$ is a mild optimum over $1$
and $8$; and prefix training costs $0.03$ within cell while slightly
improving cross-domain transfer. Meta-training itself is the only necessity: the untrained encoder trails every trained variant in domain (its cross-domain advantage is the domain-bound effect discussed in Section~\ref{sec:exp-generalization}). At inference the
prototype readout is robust: cosine and logistic-regression readouts
match Euclidean, $k$NN is clearly worse, and stratifying the support
draw by class is worth a few points. Finally, the prototype
\emph{distance} used in training is the one recipe choice that matters
on diverse pools: in separate paired runs, squared
\citep{snell2017prototypical} and unsquared Euclidean are equivalent
when training on a single-policy pool (in-domain $.886$ vs.\ $.887$,
mean over the three Tau2-Bench cells), but on the all-policy pool the
unsquared form is outperforms (BFCLv3 in-domain $.877$ vs.\ $.843$).
All final models use unsquared Euclidean distance.

\section{Hyperparameter Details}
\label{app:hyperparameters}
Table~\ref{tab:hyperparams-protonet} summarizes the \protonet{} encoder architecture and training configuration. Table~\ref{tab:hyperparams-grpo} lists the GRPO training setup used for all early stopping experiments.

\begin{table}[t]
\centering
\small
\begin{tabular}{l l}
\toprule
Parameter & Value \\
\midrule
\multicolumn{2}{l}{\emph{Embedding}} \\
Text encoder & \texttt{text-embedding-3-large} \\
Embedding dimension & 3072 \\
\midrule
\multicolumn{2}{l}{\emph{Encoder Architecture}} \\
Projection dimension $d'$ & 128 \\
Transformer layers & 2 \\
Attention heads & 4 \\
FFN dimension & 512 \\
Dropout & 0.3 \\
Prototype distance & unsquared Euclidean \\
\midrule
\multicolumn{2}{l}{\emph{Episodic Training}} \\
Episodes per epoch & 100 \\
Epochs & 100 \\
Support size $N_S$ (per class) & 4 \\
Query size $N_Q$ (per class) & 4 \\
Learning rate & $1 \times 10^{-4}$ \\
Weight decay & $1 \times 10^{-2}$ \\
Warmup fraction & 0.05 \\
\midrule
\multicolumn{2}{l}{\emph{Data}} \\
Rollouts per task & 50 \\
Training-pool filter & $\geq 8$ rollouts per class \\
Evaluation filter & $\geq 5$ rollouts per class \\
Cross-validation folds & 3 \\
\bottomrule
\end{tabular}
\caption{\protonet{} encoder and training hyperparameters (the final
sweep configuration; the design study of Appendices~C and~F uses the
same architecture at 4{,}000 episodes).}
\label{tab:hyperparams-protonet}
\end{table}

\begin{table}[t]
\centering
\small
\begin{tabular}{l l}
\toprule
Parameter & Value \\
\midrule
\multicolumn{2}{l}{\emph{Model}} \\
Base model & Qwen3-8B \\
LoRA rank & 32 \\
\midrule
\multicolumn{2}{l}{\emph{GRPO}} \\
Learning rate & $2 \times 10^{-5}$ \\
Loss function & PPO (clip 0.8--1.2) \\
Substeps per batch & 2 \\
Group size & 16 rollouts \\
Batch size & 20 tasks \\
Training batches & 50 \\
\midrule
\multicolumn{2}{l}{\emph{Rollout Generation}} \\
Max tokens per turn & 2{,}048 \\
Temperature & 1.0 \\
Max turns & 100 \\
User simulator & \usersim \\
\midrule
\multicolumn{2}{l}{\emph{Early Stopping}} \\
Controller & prefix-trained \protonet{} \\
Threshold $\theta$ & $\{0.65, 0.7, 0.8\}$ \\
Forced completions $m_c$ & $\{0, 4\}$ per group \\
Min steps before stopping & 1 \\
Rolling support window & 3 batches \\
Min class support & 16 \\
Embedding model & \texttt{text-embedding-3-large} \\
\midrule
\multicolumn{2}{l}{\emph{Evaluation}} \\
Eval frequency & every 5 batches \\
Eval group size & 3 rollouts \\
Task split & official Tau2-retail, 74 train / 40 test \\
Seeds & 3 \\
\bottomrule
\end{tabular}
\caption{GRPO training and early stopping configuration.}
\label{tab:hyperparams-grpo}
\end{table}

\section{Support Set Full Method Results}
\label{app:support-set-full}

Figure~\ref{fig:fewshot} compares \protonet{} against the strongest
per-task baseline at each support size; Table~\ref{tab:fewshot-full}
reports every method. The protocol is identical across methods: for
each held-out test task and each $k \in \{1, 2, 4, 8\}$, the same five
seeded draws of $k$ rollouts per class are used --- the per-task
discriminators (NC and RBF on the pooled representation, fastText
$n{=}1$ on state-ID sequences) are fit on the draw, while \protonet{}
(per-benchmark specialist, averaged over 3 training seeds) uses it as
its support set with no fitting. Values are mean per-task \auroc{},
averaged over the benchmark's three policy cells. Higher-order fastText
variants ($n \in \{2, 5\}$) are omitted: at these support sizes they
have too few sequences to fit n-gram statistics and fall below
$n{=}1$.

\begin{table}[t]
\centering
\small
\setlength{\tabcolsep}{5pt}
\begin{tabular}{l c cccc}
\toprule
& $k$ & \protonet{} & NC & RBF & FT $n{=}1$ \\
\midrule
\multirow{4}{*}{Tau2-Bench}
& 1 & \textbf{.871} & .743 & .751 & .705 \\
& 2 & \textbf{.903} & .783 & .795 & .751 \\
& 4 & \textbf{.918} & .825 & .839 & .799 \\
& 8 & \textbf{.926} & .855 & .874 & .857 \\
\midrule
\multirow{4}{*}{BFCLv3}
& 1 & \textbf{.840} & .724 & .729 & .691 \\
& 2 & \textbf{.856} & .766 & .773 & .755 \\
& 4 & \textbf{.876} & .801 & .811 & .804 \\
& 8 & \textbf{.885} & .830 & .840 & .852 \\
\midrule
\multirow{4}{*}{AgentClinic}
& 1 & \textbf{.713} & .645 & .651 & .572 \\
& 2 & \textbf{.763} & .696 & .702 & .595 \\
& 4 & \textbf{.784} & .722 & .728 & .643 \\
& 8 & \textbf{.805} & .750 & .765 & .705 \\
\bottomrule
\end{tabular}
\caption{Few-shot outcome discrimination on held-out test tasks
(\auroc{}, mean over tasks and the benchmark's three policy cells;
identical seeded support draws across methods, five draws per task and
$k$; AgentClinic restricted to tool-reliant scenarios). Per-task
discriminators are fit on the $k$-per-class draw; \protonet{} uses it
as its support set. Best per row in bold.}
\label{tab:fewshot-full}
\end{table}

\protonet{} leads every method at every support size on every
benchmark. The margin is widest at $k{=}1$, where per-task fitting has
almost nothing to fit on, and narrows as $k$ grows. At $k{=}8$ the
strongest baseline closes to within $0.03$--$0.05$, consistent with
Section~\ref{sec:exp-sep}'s finding that given enough supervised data the naive discriminators are able to approach \protonet{}'s performance.

\section{AgentClinic Analysis}
\label{app:agentclinic}
AgentClinic is the consistent low performer across
Table~\ref{tab:separability}, the few-shot sweep, and the
generalization grids, for per-task discriminators and \protonet{}
alike. This appendix locates the cause in the benchmark's structure:
its outcome often does not flow through the observable
action--observation stream.

Unlike Tau2-Bench and BFCLv3, where success requires a specific
sequence of correct API calls whose effects are visible in the
observations, an AgentClinic rollout succeeds if the final diagnosis is
correct, and that diagnosis can frequently be produced from the
model's internal medical knowledge, with the patient dialogue acting as
a thin retrieval layer. We perform a no-tools ablation to quantify this. Every mixed-outcome scenario is rerun with a doctor prompt stripped of the ability to order tests.  Each scenario is labeled tool-reliant if its success rate drops by more than $0.1$ when tools are removed, tested one-sided against that $0.1$ margin (50 with-tools rollouts against 25 or no-tools rollouts per scenario), under false-discovery-rate control at 5\% within each model. Most scenarios show no confirmed drop: 83\% for \oaimodel{} (124 of 150), 72\% for \qwenbig{} (84 of 117), and 62\% for \dsmodel{} (69 of 112). On these scenarios the trajectory's actions and observations are largely epiphenomenal to the outcome, so no representation over them should be expected to separate outcomes well.

\subsection{Is the Outcome Determined by the Trajectory?}
A discriminator over actions and observations can only predict what the
trajectory determines. The tool-reliance analysis above shows that much
of AgentClinic's outcome variance comes from knowledge the trajectory
never exposes. In this subsection, for a rollout that ends in a diagnosis, we reconstruct the doctor's context up to --- but excluding --- its final diagnosis turn, and resample that
turn ten times from the same policy at generation temperature, and
grade each resample using the environment's grader. If the outcome is a function of the trajectory, resampled outcomes should agree with the original; if it is decided at the final synthesis step, they should flip. We run this on 20 mixed-outcome scenarios per policy, 10 rollouts each, for a total of $2{,}000$ resampled diagnoses per policy.

\begin{table}[t]
\centering
\small
\setlength{\tabcolsep}{5pt}
\begin{tabular}{l ccc}
\toprule
Policy & Cons. & Ambig. & \auroc{} ceil. \\
\midrule
\dsmodel      & .48 & .28 & .918 \\
\qwenbig      & .61 & .17 & .948 \\
\oaimodel     & .40 & .34 & .887 \\
GPT-5.6-Terra & .95 & .03 & .994 \\
\bottomrule
\end{tabular}
\caption{Final-diagnosis resampling on AgentClinic (20 scenarios
$\times$ 10 rollouts $\times$ 10 resamples per policy. With $\hat{p}$ for the fraction of a rollout's resamples graded correct: \emph{Consistent} = fraction of rollouts with $\hat{p} \in \{0, 1\}$; \emph{Ambiguous} = fraction with $0.2 < \hat{p} < 0.8$; \emph{\auroc{} ceiling} = expected per-task \auroc{} of an oracle that scores each rollout by its own $\hat{p}$, under outcomes drawn from those same $\hat{p}$ --- the best any trajectory representation could achieve if it recovered the prefix's outcome distribution exactly.}
\label{tab:ac-resample}
\end{table}

Outcomes are mostly, but not entirely, determined by the trajectory.
For the three policies of our main experiments, only $40$--$61\%$ of
rollouts resample deterministically, and $17$--$34\%$ are genuinely
ambiguous: the same diagnostic history yields both correct and
incorrect diagnoses depending on the sampling draw. This stochasticity
is irreducible for any representation of the trajectory, and it caps
achievable \auroc{} at $.89$--$.95$ --- a ceiling with no counterpart
in Tau2-Bench or BFCLv3, where the environment's state rather than a
final free-text synthesis determines the reward. The fourth policy
locates the effect in the policy rather than the benchmark: GPT-5.6-Terra resamples deterministically on $95\%$ of rollouts, with a
ceiling of $.99$ --- its diagnosis is essentially fixed by the
interview it conducted. On AgentClinic, then, outcome predictability
from the trajectory is bounded by how reliably the policy converts its
own gathered evidence into a diagnosis, and for weaker policies that
conversion is itself a coin flip on a substantial minority of
rollouts.

\begin{figure}[!b]
    \centering
    \includegraphics[width=0.9\columnwidth]{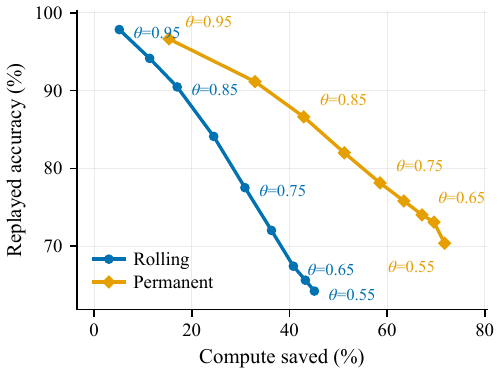}
    \caption{Simulated early-stopping frontier: replayed accuracy against compute saved, sweeping the stopping threshold $\theta$ from $0.95$ (conservative, few stops) to $0.55$ (aggressive) in steps of $0.05$. Replayed over the completed rollouts of three baseline GRPO seeds, for support accrued during the run (rolling) and pre-collected per task (permanent).}
    \label{fig:strictness}
\end{figure}

\section{Cross-Benchmark Generalization}
\label{app:cross-benchmark}

Table~\ref{tab:gen-benchmarks} repeats the analysis of
Section~\ref{sec:exp-generalization} along the benchmark axis: training pools
contain all three policies' rollouts from one, two, or all three benchmarks,
and each pool is evaluated on the held-out test tasks of every benchmark,
aggregated over the three policies and 3 training seeds. As a reference, the
final row reports the untrained encoder: the same architecture with randomly
initialized projection and transformer weights, evaluated under the identical
support protocol.

\begin{table}[t]
\centering
\small
\setlength{\tabcolsep}{5pt}
\begin{tabular}{l ccc}
\toprule
& \multicolumn{3}{c}{\textbf{Evaluated on}} \\
\cmidrule(lr){2-4}
\textbf{Trained on} & \textsc{Tau2} & \textsc{BFCLv3} & \textsc{AgentClinic} \\
\midrule
Tau2-Bench          & .923        & \ood{.745}  & \ood{.619} \\
BFCLv3              & \ood{.763}  & .879        & \ood{.617} \\
AgentClinic         & \ood{.796}  & \ood{.777}  & .793 \\
\addlinespace[2pt]
Tau2+BFCL           & .919        & .872        & \ood{.623} \\
Tau2+AC             & .915        & \ood{.763}  & .793 \\
BFCL+AC             & \ood{.816}  & .869        & .798 \\
\addlinespace[2pt]
All benchmarks      & .908        & .870        & .789 \\
\midrule
\emph{Untrained encoder} & \emph{.829} & \emph{.807} & \emph{.678} \\
\bottomrule
\end{tabular}
\caption{Cross-benchmark generalization: held-out-task \auroc{} by
training-pool benchmark composition (rows; all three policies, aggregated over
policies and 3 seeds) and evaluated benchmark (columns). Shaded cells are
out-of-distribution (evaluated benchmark not in the pool). The untrained
encoder is the same architecture with random weights under the identical
support protocol.}
\label{tab:gen-benchmarks}
\end{table}

Three observations. First, benchmarks must be represented in meta-training:
every out-of-distribution cell falls below its column's ID values by a wide
margin ($0.10$--$0.26$ \auroc{}), and adding a second foreign benchmark to the
pool barely moves it (e.g.\ AgentClinic OOD: $.617$--$.619$ with one foreign
benchmark, $.623$ with two). Second, OOD performance sits \emph{at or below
the untrained encoder} in every case. Meta-training on foreign domains does
not merely fail to help; it overwrites raw AOS structure that a random
projection preserves, so a deployed discriminator should not be pointed at an
unseen domain on the expectation of graceful degradation. Third, benchmark
diversity carries no penalty in domain: the all-benchmark generalist matches
the per-benchmark specialists on every benchmark ($.908$/$.870$/$.789$ vs.\
$.923$/$.879$/$.793$), so a single model can cover all domains represented in
its training pool. Together with the cross-model results of
Section~\ref{sec:exp-generalization}, this locates the learned space as
policy-general but domain-bound: coverage of the target domain is necessary,
and, because support sets are byproducts of training (Section~\ref{sec:preliminaries}),
essentially free to obtain.

\section{FLOPs Accounting}
\label{app:flops}
All compute in Section~\ref{sec:application} is reported as estimated
FLOPs for the policy model ($N = 8.19 \times 10^9$ parameters),
counting tokens through the three stages the controller affects:
\begin{equation*}
\mathrm{FLOPs} \;=\; 2N\,T_{\mathrm{gen}} \;+\; 6N\,T_{\mathrm{train}}
\;+\; 2N_{\mathrm{emb}}\,T_{\mathrm{emb}},
\end{equation*}
with $2N$ per token in a forward pass and $6N$ per token in a training
step (forward plus backward), for the policy model
$N = 8.19 \times 10^9$. The controller's encoder has an undisclosed
parameter count $N_{\mathrm{emb}}$, so its term cannot be measured; we
report the two ends of the range instead. Setting
$N_{\mathrm{emb}} = 0$ gives the savings excluding controller overhead;
setting $N_{\mathrm{emb}} = N$ --- charging every embedded token as if
the encoder were as large as the \qwensmall{} policy --- gives a
conservative upper bound on that overhead, and is the ``net'' column of
Table~\ref{tab:flops-detail} and of Table~\ref{tab:grpo}.

\paragraph{Token counts.} Generation tokens $T_{\mathrm{gen}}$ ---
prefill plus decode, summed over every rollout in the matched 50-batch
window --- are exact, recorded per rollout in the training logs.
Training tokens $T_{\mathrm{train}}$ cover the trajectories that
survive GRPO's group filter: groups whose rollouts all receive the same
reward carry no advantage signal and are dropped, so a batch trains on
the remaining $8$--$20$ groups of 16. We count these tokens directly with the policy's own tokenizer. Controller tokens $T_{\mathrm{emb}}$, the step texts embedded while scoring in-progress rollouts are recovered the same way.

\paragraph{Where the compute goes.} Generation accounts for $89\%$ of
baseline FLOPs and training the remaining $11\%$, which is why
terminating rollouts early translates almost directly into total
savings. The controller's footprint is best stated in the currency that
does not depend on $N_{\mathrm{emb}}$: embedded tokens are
$2.4$--$3.0\%$ of generated tokens in every configuration. Charged at
policy parity, this costs $0.8$--$2.3$ points of savings.

\begin{table}[t]
\centering
\small
\setlength{\tabcolsep}{5pt}
\begin{tabular}{l ccc c}
\toprule
& \multicolumn{3}{c}{Tokens (G)} & FLOPs \\
\cmidrule(lr){2-4}
Configuration & gen & train & emb & saved (\%) \\
\midrule
GRPO baseline                      & 2.49 & .108 & ---   & ---            \\
rolling $\theta{=}.8$, $m_c{=}4$   & 2.15 & .103 & .065 & $10.1 \pm 1.4$ \\
rolling $\theta{=}.7$, $m_c{=}4$   & 1.80 & .082 & .050 & $25.3 \pm 3.6$ \\
rolling $\theta{=}.7$, $m_c{=}0$   & 1.55 & .075 & .042 & $35.5 \pm 2.2$ \\
rolling $\theta{=}.65$, $m_c{=}0$  & 1.40 & .057 & .035 & $43.0 \pm 4.2$ \\
permanent $\theta{=}.8$, $m_c{=}0$ & 1.25 & .067 & .034 & $47.3 \pm 1.8$ \\
permanent $\theta{=}.7$, $m_c{=}0$ & 0.85 & .055 & .022 & $63.0 \pm 2.0$ \\
\bottomrule
\end{tabular}
\caption{Token budget and resulting FLOPs savings per configuration
(mean over 3 seeds, batches $<50$). Tokens in billions; savings charge
the controller at policy parity ($N_{\mathrm{emb}} = N$) and match
Table~\ref{tab:grpo}. Baseline total: $46.0 \pm 0.1$ EFLOPs.}
\label{tab:flops-detail}
\end{table}

\section{Separability on Shared Tasks}
\label{app:sep-shared-tasks}
Table~\ref{tab:separability} evaluates each (benchmark, policy) cell on
its own held-out tasks, so $N$ varies across policies: a task enters a
cell's evaluation set only if that policy produced at least five
rollouts of each outcome class on it, and weaker policies fail
different tasks than stronger ones. Differences down a benchmark's rows
therefore confound the policy with the task set. This appendix repeats
the comparison on the held-out tasks that pass the
filter under \emph{all three} policies, so that every row of a
benchmark block is scored on identical tasks.

\begin{table*}[t]
\centering
\small
\setlength{\tabcolsep}{3.5pt}
\begin{tabular}{ll ccccc c}
\toprule
& & \multicolumn{2}{c}{Pooled} & \multicolumn{3}{c}{Clustered} & Learned \\
\cmidrule(lr){3-4}\cmidrule(lr){5-7}\cmidrule(l){8-8}
Dataset & Model & NC & RBF & $n{=}1$ & $n{=}2$ & $n{=}5$ & \protonet{} \\
\midrule
\multirow{3}{*}{Tau2-Bench}
& DS-V3.2$^\star$    & .926 & .946 & .847 & .859 & .839 & \textbf{.970} \\
& Qwen3-80B$^\dagger$ & .858 & .917 & \textbf{.956} & .952 & .920 & .937 \\
& GPT-5-mini$^\ddagger$ & .882 & .933 & \textbf{.950} & .943 & .901 & \textbf{.950} \\
\midrule
\multirow{3}{*}{BFCLv3}
& DS-V3.2$^\star$    & .802 & .823 & \textbf{.896} & .894 & .867 & .855 \\
& Qwen3-80B$^\dagger$ & .938 & \textbf{.960} & .940 & .938 & .915 & .931 \\
& GPT-5-mini$^\ddagger$ & .858 & .930 & .894 & \textbf{.936} & .934 & .861 \\
\midrule
\multirow{3}{*}{AgentClinic}
& DS-V3.2$^\star$    & .750 & \textbf{.786} & .654 & .636 & .621 & .705 \\
& Qwen3-80B$^\dagger$ & .730 & .867 & .738 & .693 & .609 & \textbf{.893} \\
& GPT-5-mini$^\ddagger$ & .800 & .807 & .776 & .728 & .683 & \textbf{.819} \\
\bottomrule
\end{tabular}
\caption{Outcome separability on the shared held-out tasks of each
benchmark (\auroc{}, mean over folds; protocol identical to
Table~\ref{tab:separability}). Shared task counts are $N{=}12$
(Tau2-Bench), $N{=}12$ (BFCLv3), and $N{=}4$ (AgentClinic); within a
benchmark, all three policy rows are scored on exactly these tasks.
Best per row in bold. $^\star$deepseek-v3.2;
$^\dagger$qwen3-next-80b-a3b-thinking; $^\ddagger$gpt-5-mini-2025-08-07.}
\label{tab:sep-shared}
\end{table*}

The overall picture is unchanged: all three representation families
separate outcomes well above chance on Tau2-Bench and BFCLv3 and less
well on AgentClinic, and no family dominates uniformly. What the
matched comparison adds is that this ordering does not depend on which
tasks each policy happened to survive --- the cross-policy differences
within a benchmark are properties of the policies, not of differing
evaluation sets.

The shared sets are small, and deliberately so: the intersection keeps
only tasks on which \emph{every} policy produces both outcomes, which
excludes tasks that are reliably solved by the strongest policy or
reliably failed by the weakest. This has three consequences. Sampling
noise is large --- with $N{=}12$ a single task shifts a cell mean by
several points, and per-cell differences below roughly $0.05$ should
not be read as real. AgentClinic is the extreme case at
$N{=}4$ as its evaluation sets are each policy's own
tool-reliant scenarios (Appendix~\ref{app:exp-setup-details}), which
overlap only slightly across policies. We therefore treat this table as a consistency check on Table~\ref{tab:separability} rather than as an independent measurement.

\section{GRPO Additional Details}
\label{app:gpro-details}
This appendix details the rollout controller's support machinery and the early stopping simulation.

\subsection{Support Sets}

A task's support set is what turns \protonet{} from an encoder into a
discriminator: prototypes are formed from labeled same-task rollouts,
and a task can only be scored once it has at least two support rollouts
covering both outcome classes. Support is keyed per task, so coverage
accrues independently for each task the scheduler visits.

\paragraph{Rolling support.} The self-contained configuration uses only
rollouts the current run has already completed. Within a batch,
completed rollouts are \emph{staged} rather than added immediately, and
committed once the batch finishes, so every rollout in a batch is
scored against the same support set and the controller cannot chase its
own within-batch output. A task therefore cannot be scored until at
least one batch containing it has completed with both outcomes present.
Because tasks rotate through batches, support is capped by a rolling
window of the three most recent batches per task, with one exception:
if applying the window would leave fewer than 16 rollouts of either
class, the most recent evicted rollouts of the deficient class are
retained. This keeps prototypes recent without letting them collapse
onto a single class as the policy improves and failures become rare.

\paragraph{Permanent support.} The alternative pre-collects 16 rollouts
per task from the corpus of Section~\ref{sec:experiments} and loads
their step embeddings before training begins. These are never evicted,
so every task is scoreable from the first batch, and the controller
pays no cold-start period. Permanent and rolling support are not
exclusive: when both are present the controller concatenates them,
so a permanent-support run also accumulates on-policy rollouts as
training proceeds. The permanent rollouts come from a different policy
than the one being trained (\qwenbig{} rather than \qwensmall{}), which
makes this configuration a live test of the cross-model generalization
of Section~\ref{sec:exp-generalization}.

\paragraph{Forced completions.} Within each group of 16 rollouts, the
first $m_c$ are exempt from early stopping and always run to
termination. This guarantees that every group contributes at least
$m_c$ true environment rewards to its advantage estimate, and under
rolling support, that fresh labeled rollouts keep arriving even when
the controller is stopping aggressively.

\subsection{Simulated Early Stopping}
Live GRPO runs are expensive, so we selected the configurations of
Section~\ref{sec:application} by first \emph{simulating} early
stopping: replaying the prefix-trained \protonet{} over the completed
rollouts of three baseline GRPO seeds. Each rollout is scored after
every action--observation pair exactly as in deployment and stops at
the first step crossing the threshold, taking the predicted outcome as
its reward. Because the full rollout and its true reward are known,
the replay measures what a live run cannot: the accuracy of the
stopped predictions, the compute that would have been saved, and the
bias that predicted rewards inject into the group's reward signal.
Figure~\ref{fig:strictness} sweeps $\theta$ from $0.55$ to $0.95$ for
the two support strategies --- \emph{rolling}, accrued from the
current run's completed rollouts, and \emph{permanent}, pre-collected
per task and available from the first batch.

Two observations shaped the live sweep. First, \textbf{permanent
support dominates rolling across the entire frontier}: at matched
replayed accuracy it saves $12$--$34$ additional percentage points of
compute, with the gap widening as the threshold loosens (at $84\%$
accuracy, $47.5\%$ vs.\ $24.5\%$ saved). Rolling support cannot score a task until that task has accumulated completed rollouts, and those rollouts are the expensive ones. Even at the most conservative threshold, where rolling saves only $5.2\%$, permanent saves $15.3\%$ at comparable accuracy. Second, \textbf{reward bias is small and shrinks with $\theta$}: at $\theta{=}0.8$ predicted rewards understate the group mean by $0.04$ (rolling) and $0.04$ (permanent), with mean absolute error $0.16$ and $0.18$, falling to $0.01$ (rolling) and $0.03$ (permanent) by $\theta{=}0.9$. This is what licensed running the live configurations without a correction term, and it motivated retaining a floor of forced completions ($m_c{=}4$) in the deployed configuration, where the bias is smallest but the support set is coldest.

\end{document}